# Memory-updated-based Framework for 100% Reliable Flexible Flat Cables Insertion

Zhengrong LING[1], Xiong YANG[1], Dong GUO[2], Hongyuan CHANG[1], Tieshan ZHANG[1], Ruijia ZHANG[1], Yajing SHEN[1,3*]

[1] Department of Electronic and Computer Engineering, The Hong Kong University of Science and Technology, Hong Kong 999077, China

[2] Department of Biomedical Engineering, City University of Hong Kong, Hong Kong 999077, China

[3] Center for Smart Manufacturing, The Hong Kong University of Science and Technology, Hong Kong 999077, China

[*]Corresponding author(s): eeyajing@ust.hk

Contributing authors: zlingab@connect.ust.hk; eexiongyang@ust.hk; dongguo3-c@my.cityu.edu.hk; hchangac@connect.ust.hk; eetieshan@ust.hk; rzhangdk@connect.ust.hk.

## Abstract

Automatic assembly lines have increasingly replaced human labor in various tasks; however, the automation of Flexible Flat Cable (FFC) insertion remains unrealized due to its high requirement for effective feedback and dynamic operation, limiting approximately 11% of global industrial capacity. Despite lots of approaches, like vision-based tactile sensors and reinforcement learning, having been proposed, the implementation of human-like high-reliable insertion (i.e., with a 100% success rate in completed insertion) remains a big challenge. Drawing inspiration from human behavior in FFC insertion, which involves sensing three-dimensional forces, translating them into physical concepts, and continuously improving estimates, we propose a novel framework. This framework includes a sensing module for collecting three-dimensional tactile data, a perception module for interpreting this data into meaningful physical signals, and a memory module based on Bayesian theory for reliability estimation and control. This strategy enables the robot to accurately assess its physical state and generate reliable status estimations and corrective actions. Experimental results demonstrate that the robot using this framework can detect alignment errors of 0.5 mm with an accuracy of 97.92% and then achieve a 100% success rate in all completed tests after a few iterations. This work addresses the challenges of unreliable perception and control in complex insertion tasks, highlighting the path toward the development of fully automated production lines.

## 1 Introduction

The automation of assembly lines has long been a key objective for enhancing productivity, precision, consistency, and quality control in industry. In recent decades, advancements in robotics have facilitated the automation of various repetitive tasks—such as welding, painting, and packaging—at high speeds with consistent accuracy[1–4]. However, the automatic insertion of flexible flat cables (FFC), which impacts approximately 11% of global industrial capacity, presents a significant challenge. This is largely due to stringent requirements for robotic systems in both hardware and algorithms, particularly in accurately identifying alignment errors and implementing effective manipulation strategies[5–7].

The most direct method for recognizing the positions of pegs and holes, as well as identifying alignment errors, is vision feedback. This approach has been widely adopted in robotic automation tasks, enabling applications such as coil-cylinder assembly and gear mating[8,9]. However, in FFC insertion, cameras are often obstructed by circuit components, leading to detection failure[7]. Furthermore, vision systems frequently struggle to detect small alignment errors accurately due to limited camera resolution and the small size of the components[10]. Force sensing provides another valuable means of reflecting contact information, a method primarily employed by humans during FFC insertion. Among various force feedback approaches, soft tactile sensors mounted on end effectors best replicate the tactile functionality of fingertip sensing, enabling direct measurement of contact force with workpieces[11–13]. However, existing sensors often lack the capability to obtain precise three-dimensional force data[14,15], which constrains the acquisition of contact status in FFC insertion. Some tactile sensors, such as GelSight[16,17] and Finger Vision[18], can decode three-dimensional contact forces by tracking the motion of markers, demonstrating significant potential for plug-in tasks[11,19]. Nonetheless, they still struggle to deliver high-resolution force information due to limitations in image resolution and complex estimation models.

In addition to the hardware challenges, the algorithms for automatic FFC insertion remain in their infancy. Raw sensory data does not directly reflect the insertion status, necessitating solutions for feature extraction from tactile data and the implementation of reliable insertion strategies. Traditionally, this has been achieved using physical analytical models, such as the three-point contact model[20] and the cylindrical peg-in-hole model[21]. However, constructing these models is complex, and incomplete or inaccurate parameters can lead to erroneous estimates of contact forces. Recently, learning-based methods have emerged to estimate contact conditions, providing a powerful means to convert abstract tactile data into lower-dimensional feature vectors[22–24]. These features usually have no physical meaning, forcing the use of data-driven controllers like Reinforcement Learning, which is difficult to apply in the industry due to issues such as sampling efficiency[25]. To date, the existing methods can only extract physically meaningful features from image tactile data, resulting in the absence of force information and preventing precise estimations[26]. A method to convert the force tactile data to physically meaningful signals is urgently needed.

Moreover, reliability is a frequently overlooked aspect of current research. Given that failed insertions can incur significant costs, industrial processes prioritize reliability even above the efficiency gained through automation. While assembly line workers can often ensure a reliable final product – even if not succeeding on the first attempt – current automated FFC insertion approaches, despite achieving success rates as high as 97%[22], still have the potential for errors due to perceptual uncertainty. Some methods, such as the Backprop Kalman Filter[27,28] and factor graph approaches[29], utilize filters to combine data from multiple iterations, helping to reduce uncertainty. However, these methods assume uncertainty follows a Gaussian distribution, which lacks evidence for effectively modeling neural network outputs and may worsen estimations. Achieving 100% success in completed FFC insertion, comparable to human performance, remains a significant challenge.

This work presents a memory-updated-based framework for FFC insertion that achieves a reliable success rate of 100%. The framework includes a three-dimensional force sensation module, a physically meaningful perception module, and a memory module with reliability control capabilities. Within this framework, tactile data are collected and transformed into perception signals. The contact status estimations are stored in the memory module and are updated by iterations, enabling the robot to estimate and control the insertion reliability. The results indicate a success rate of 100% can be achieved in the completed insertions after a few iterations. This work addresses the challenges of unreliable perception and control in complex insertion tasks and provides a solution for the realization of fully automated production lines.

## 2 Result

### 2.1 Overview of the Memory-updated-based FFC Insertion Framework

As illustrated in Fig. 1A, humans not only sense three-dimensional forces through tactile receptors but also encode nerve stimulation into meaningful signals, such as success, left-offset, and right-offset. More importantly, unlike each insertion iteration that occurs without prior experience, humans retain possible contact statuses based on past perceptions. By utilizing new perception information, they can refine and update these contact statuses, improving estimation and action. Inspired by human capabilities, we propose a highly reliable FFC insertion framework, which comprises a sensation module, a perception module, a memory module, and a robotic arm equipped with grippers, as shown in Fig. 1B.

The hardware, as depicted in Fig. 1B, facilitates insertion actions and tactile sensing. During insertion, the robot executes the insertion based on the action command $u_i$, and detects the three-dimensional contact forces $F_i(1), F_i(2), ..., F_i(T)$ which are subsequently converted into digital tactile signals $d_i(t)=\{x_i(t), y_i(t), z_i(t)\}$, where $t = 1,2,...,T$, by the tactile sensor. These tactile signals are then transformed into perception signals $z_i$ through the perception module. Lastly, the memory module updates the estimation of contact status based on the $z_i$ and historical experience. In the meantime, the probabilities of all the contact statuses are calculated as reliability $\hat{y}_i$, which are used to generate the optimal action command $u_i$ until the reliability of successful insertion meets the target.

### 2.2 Three-dimensional High-Resolution Tactile Information by the Sensation Modules

As shown in Fig. 2A, the tactile sensor with three-dimensional force decoupling capabilities was integrated into the grippers. This sensor measures applied forces by detecting changes in the internal magnetic field, as detailed in Supplementary Material S1. The magnetic field is sinusoidal and arranged in an annular configuration, enabling the sensor to indicate three-dimensional forces through the output of tactile data. This allows it to detect both normal and omnidirectional shear forces. Force components in the X, Y, and Z directions correspond to changes in the sensor's X, Y, and Z values, respectively. Fig. 2(B-D) illustrates how the tactile data reflects the insertion status—success, left-offset, and right-offset—during the FFC insertion process. Specifically, when the FFC is successfully inserted, the tactile data in the X direction shows a prominent peak, with minimal changes in the other axes, as the contact forces primarily act upward along the X-axis. Conversely, unsuccessful insertion results in a decrease in Y-axis force due to right alignment errors and an increase due to left alignment errors.

Given that FFC insertion is a continuous process, the dynamic changes in three-dimensional forces provide critical signals that indicate the status of the insertion. To investigate these changes, we collected tactile data, as shown in Fig. 3. Initially, the FFC is adjusted to the reference position, either at the left edge ($P_L$) or the right edge ($P_R$), where is the most edge position that can be successfully inserted. The robot arm then moves a distance of $X_1$ to the offset position and performs the FFC insertion, during which a sample of tactile data is recorded. Ultimately, we gathered 385 series of tactile data from the insertion process at intervals of 0.05 mm within the range of $[P_L - 1.50, P_R + 1.50]$ mm.

The tactile data, represented as three-dimensional time series trajectories, as shown in Fig. 3B, reveals distinct trends during insertion, as indicated by the black arrows. Successful insertion data, shown in purple, exhibits minimal changes in the Y and Z directions, while X-axis values initially rise and then fall due to the varying friction. Data from left or right offsets trend positively along the X-axis, with left offsets increasing Y-axis values and right offsets decreasing them, highlighting the influence of offset direction on tactile data trends. As the offset magnitude increases, Y-axis changes initially rise and then fall, while Z-axis changes continue to increase, creating a curved surface distribution in three-dimensional space. Thus, the tactile data effectively captures both the direction and magnitude of alignment error, reflecting the contact status of the FFC.

**2.3 Physically Meaningful Signals Extraction by the Perception Module**

To interpret tactile data into physically meaningful signals, we developed an encoding method to describe the FFC position and implemented a neural network model to convert the tactile data. As shown in Fig. 4A, the connector is manufactured with a clearance ($\delta$) to facilitate the insertion of the FFC into the socket. Consequently, not every insertion or alignment error requires a unique adjustment distance; instead, a group of adjacent positions, represented by the colored regions in Fig. 4A, can share the same adjustment command to return to an insertable position. Thus, we propose using several position regions, rather than exact position values, to describe the status of the FFC.

To fully describe all statuses of the FFC, we designed three classes of error regions, as shown in Fig. 4B: the middle region, the left error region, and the right error region, denoted as M, L, and R, respectively. The middle region represents the continuous set of positions where the FFC can be inserted. The left and right regions are subdivided into L1, L2, ..., Ln and R1, R2, ..., Rn based on the magnitude of the deviation from the middle region, where $n$ indicates the deviation distance $\Delta = n\delta$.

As illustrated in Figs. 3, the dynamics of the three-dimensional forces reflect the magnitude and direction of the alignment error. To capture these features, we implement a Temporal Convolutional Network (TCN) block[31], as illustrated in Fig. 4C. The TCN captures features over short periods with small kernels and identifies macro trends through dilated kernels, enabling it to effectively obtain both local and global trends in the tactile data, thereby enhancing status estimation performance.

In the experiment, $\delta$ is set to 0.5 mm, and we encode the contact status of the FFC using seven regions, assuming that the previous visual positioning limits the insertion or alignment error to 1.5 mm, i.e. $e_m < 1.5$mm. We collected 530 validated samples to train and test the perception module. Out of these, 385 samples were designated for testing, with 55 samples allocated for each region. As shown in Table 1, the perception module achieved an accuracy of over 96.36% for each contact status. Furthermore, all incorrect estimations are clustered near the true status, and the total estimation error for each status does not exceed 3.64%. Overall, the perception module achieves an accuracy of 97.92% in estimating the contact status of the FFC, providing accurate and meaningful physical signals.

## 2.4 Reliability Estimation and Control by the Memory Module

In conventional methods, while the estimation of contact status may achieve high accuracy, there remains a probability of incorrect estimations that can lead to insertion failure. For example, the system may erroneously conclude that the insertion was successful when it has actually failed. In our approach, we introduce a memory module to maintain a probability distribution of the contact status, reflecting the reliability of these estimates. We utilize the latest perceptual signals to update this distribution of contact status based on Bayes' Theorem. At each iteration, only the most reliable status is accepted to generate adjustment commands leading to successful completed insertion.

As illustrated in Section 2.3, the FFC status $s$ can be labeled according to the error region in which the FFC is located. We digitize these statuses in order using $s \in S = \{-n, \ldots, -2, -1, 0, 1, 2, \ldots, n\}$, where $n$ represents the status when the alignment error is $n\delta$. The value of $s$ represents the magnitude of the alignment error, while the sign of $s$ indicates left (positive) and right (negative) alignment errors. The FFC may be in one of these statuses, and their probabilities are represented by a reliability distribution $P(s)$. Each time the memory module outputs the estimated status $\hat{s}_i$, its reliability is $\gamma_i = P(\hat{s}_i)$.

Each iteration, such as the $i$-th iteration, begins with an adjustment action $u_i \in U = \{-n, \ldots, -2, -1, 0, 1, 2, \ldots, n\}$, where $n$ represents the movement distance of $n\delta$, and the sign of $u_i$ indicates the adjustment direction: right (positive) or left (negative). The action results in the first update of the reliability distribution from $P_i(s)$ to $\tilde{P}_i(s)$ as follows:

$$\tilde{P}_i(s) = \begin{cases} P_i(s - u_i) & if\ s - u_i \in S \\ 0 & if\ s - u_i \notin S \end{cases} \quad (1)$$

After shifting the distribution according to the adjustment action, an insertion is performed generating tactile data, which is subsequently converted to the perception signal $z_i$. Then the reliability distribution is updated again as follows:

$$P_{i+1}(s) = \eta p(z_i \mid s) \tilde{P}_i(s) \quad (2)$$

where $\eta$ is the normalization and the constant $p(z_i \mid s_i)$ is the perception probability of $z_i$ given $s_i$, which can be obtained through perception model testing.

In a typical insertion process, as illustrated in iteration 0 in Fig. 5, the initial action is set to $u_0 = 0$, updating the reliability distribution according to Eq. (1). Next, the first insertion action generates tactile data, which is converted into the first measurement $z_0$. As shown in Fig. 5, the first perception signal R1 is incorrect, as the actual status is R2; however, it will not be directly used to generate an adjustment command. Instead, it is sent to update the reliability distribution based on Eq. (2), resulting in the status with maximum reliability $\hat{s}_i = \arg\max_s P_i(s)$. This process prevents the incorrect perception from directly influencing the adjustment and considers other possible statuses. In each subsequent iteration $i$, the adjustment action is determined by $u_{i+1} = -\hat{s}_{i+1}$ to compensate for the insertion alignment error. The iteration continues until the reliability of the success status M reaches a target reliability $\gamma_{\text{target}}$. The target reliability should exceed all values in the perception distribution and be close to 1 to ensure a highly reliable final insertion.

The implementation is outlined in Algorithm 1. Given the status set, action set, and perception model with its associated probabilities, the algorithm directs the robot to position the FFC reliably. In each iteration, an adjustment action updates the reliability distribution initially. This is followed by an insertion action, collection of tactile data, and generation of perception signals. The reliability distribution is then updated based on the perception signals. The loop terminates if the target reliability is met and the contact status is M; otherwise, the robot proceeds to the next iteration.

---

**Algorithm 1** Reliability Control

---

**Require:** Initialization: Status Set $\mathbb{S}$, Perception Model: $\mathcal{F}: \mathbb{R}_{n \times 3} \to \mathbb{S}$, Perception Probability: $\mathbb{M}_{7 \times 7}$, Initial Reliability List: $P = P_0$, Initial Estimated Status: $s = \hat{s}_0 = 0$

1: **while** $\hat{s} \neq 0, \gamma \leq \gamma_{target}$ **do**
2:      $u \leftarrow -\hat{s}$
3:      Take action by shifting $u$ units & update $P$
4:      Insertion & $d \leftarrow$ Collect_Data()
5:      Perception: $z \leftarrow \mathcal{F}(d)$
6:      Obtain perception probability: $\omega \leftarrow \mathbb{M}(:, z)$
7:      $P \leftarrow$ Normalization( $\omega^\top \cdot \text{Diag}(P)$ )
8:      $\hat{s} \leftarrow \arg\max P(s), \gamma \leftarrow \max P(s)$
9: **end while**

---

**2.5 Verification of the Reliability Estimation and Control in FFC Insertion**

To validate the proposed framework, we employed a 15-pin FFC with a pin pitch of 1 mm for the insertion task. As illustrated in Fig. 6 (A and B), the gripper is equipped with a tactile sensor (AgileReach Limited), and the FFC is manipulated by a six-degree-of-freedom robot arm (PHI-140-80 from DH-Robotics). Conventional visual alignment is typically limited alignment error to the range $I_e = [P_L - 1.5, P_R + 1.5]$mm. Therefore, we established the initial offset within this range for our tests. The status of the FFC is categorized into seven regions based on an insertion clearance of $\delta = 0.5$mm. We trained the perception module and derived the perception probability matrix $p(z \mid s)$, as presented in Table 2. The target reliability $\gamma_{\text{target}}$ is set to 0.999, indicating a high level of reliability.

The insertion process, illustrated in Fig. 6C, involves moving the FFC through insertion, ascent, and translation. Initially, the FFC is offset, and tactile sensor values are calibrated to zero. The first insertion causes the FFC to move downward, but an alignment error leads to contact with the socket edge, changing the tactile values (stage ①). The robot uses this data to assess contact status and reliability, then lifts and adjusts the FFC (stages ② and ③). Another insertion follows, collecting new tactile data to reassess contact status. This process repeats until the reliability exceeds the target reliability and the insertion concludes successfully (stage ⑥).

Fig. 7 gives the detailed results of 100 insertion tests for the proposed framework. The most significant change brought by the memory module to the system is its correction of the perceptions. As shown in Fig. 7A, the memory module revised part of the outputs of the perception module. Especially, the majority of the modifications occurred in iteration 1, with 11 perceptions being adjusted. These adjustments led to notable improvements in the estimation results as shown in Fig. 7B. In the initial iteration 0, the accuracy of the perception module and the memory module are the same, while the accuracy of the memory module is always significantly higher in the following iterations. Notably, in iteration 3, the perception module achieved only 50% accuracy, whereas the memory module attained 100% accuracy.

In addition to estimating the status, the memory module also provides estimates of the reliability of that status, as illustrated in Fig. 7C. As the iterations progress, the average reliability of the incorrect estimates in each iteration fluctuates below target reliability until no incorrect estimates exist. In comparison, the average reliability of the correct estimates and the average reliability of all estimates show an overall upward trend, reaching target reliability in iteration 3. Importantly, the estimated status when the insertion stops consistently remained above 0.999, indicating strong reliability.

With high reliability in status estimation, the memory module can enhance the reliability of the completed insertion. As shown in Fig. 7D, with the memory module, the insertion status, i.e., success or failure, can be estimated with a 100% success rate in each iteration. After each iteration, the robot adjusts the cable's position based on the estimated status and conducts the insertion again until the success status is reached. Moreover, it can be seen that the average distance from the FFC to the midpoint gradually decreased when utilizing the memory module (Fig. 7E), i.e., the mean absolute error (MAE) gradually approached 0 from 1.6 initially. Consequently, as shown in Fig. 7F, the overall success rate of the 100 tests continuously increased with the iterations, achieving a 100% success rate by the end.

For comparison, we also conducted 100 insertion tests using only the sensation and perception modules, excluding the memory module. The results indicate that two of these tests failed in stopping at iteration 2 (Fig. 7D). For instance, as shown in Tables S2–S4 of the Supplementary Material, unlike the test with memory modules, when the perception module estimated a false positive successful insertion, it doesn't calculate the reliability before providing the optimal action, resulting a stop at the wrong position. Moreover, without the memory module, the average distance from the FFC to the midpoint did not continuously decrease and even exhibited new errors in iterations 2 and 4, resulting in fluctuating mean absolute error (MAE) (Fig. 7E). For instance, as detailed in Table S5 of the Supplementary Material S5, the robot estimations became trapped in a loop between -1 and 1, causing the robot to adjust left and right repeatedly, and failing to complete the insertion. Consequently, as shown in Fig. 7F, the approach without the memory module cannot successfully complete all insertions.

Overall, the solution without the memory module may fail to achieve a 100% success rate due to premature termination and constant adjustments. The proposed framework utilizes the memory module to estimate and control reliability, thereby preventing the introduction of perception errors and ensuring reliable estimations and stopping decisions. As a result, incorrect perception signals are not easily adopted, leading to a 100% success rate for all stopped insertions. This framework provides a solution that guarantees reliable completed insertion results

## 3 Discussion

High-reliability FFC assembly has long presented a significant challenge in the industry. To address this issue, we propose a comprehensive framework that mimics human sensing and memory updates. Unlike conventional methods, our approach incorporates a memory module to enhance perception reliability. As a result, we achieved a 100% success rate in the completed insertion, effectively preventing erroneous halts and continuous misestimations. The proposed framework provides a solution that enables the robot to not only finish the insertion but also ensure the reliability of the completed insertion. Note that this work focuses on the most challenging insertion process in the FFC assembly. The pre-steps, such as FFC grasping and visual alignment, and the wrong output from the sensor and robot were not considered, the failure of which also affects the final success rate. In the future work, we will try to integrate all assembly steps and systematically consider reliability, thereby achieving a high success rate for the entire assembly process.

## Data Availability

The datasets generated and/or analysed during the current study are available in the "NpjRoboticsData" repository, https://github.com/soil-code/NpjRoboticsData.git.

## Code Availability

The underlying code and training/validation datasets for this study are available in the "NpjRoboticsCode" repository and can be accessed via this link: https://github.com/soil-code/NpjRoboticsCode.git.

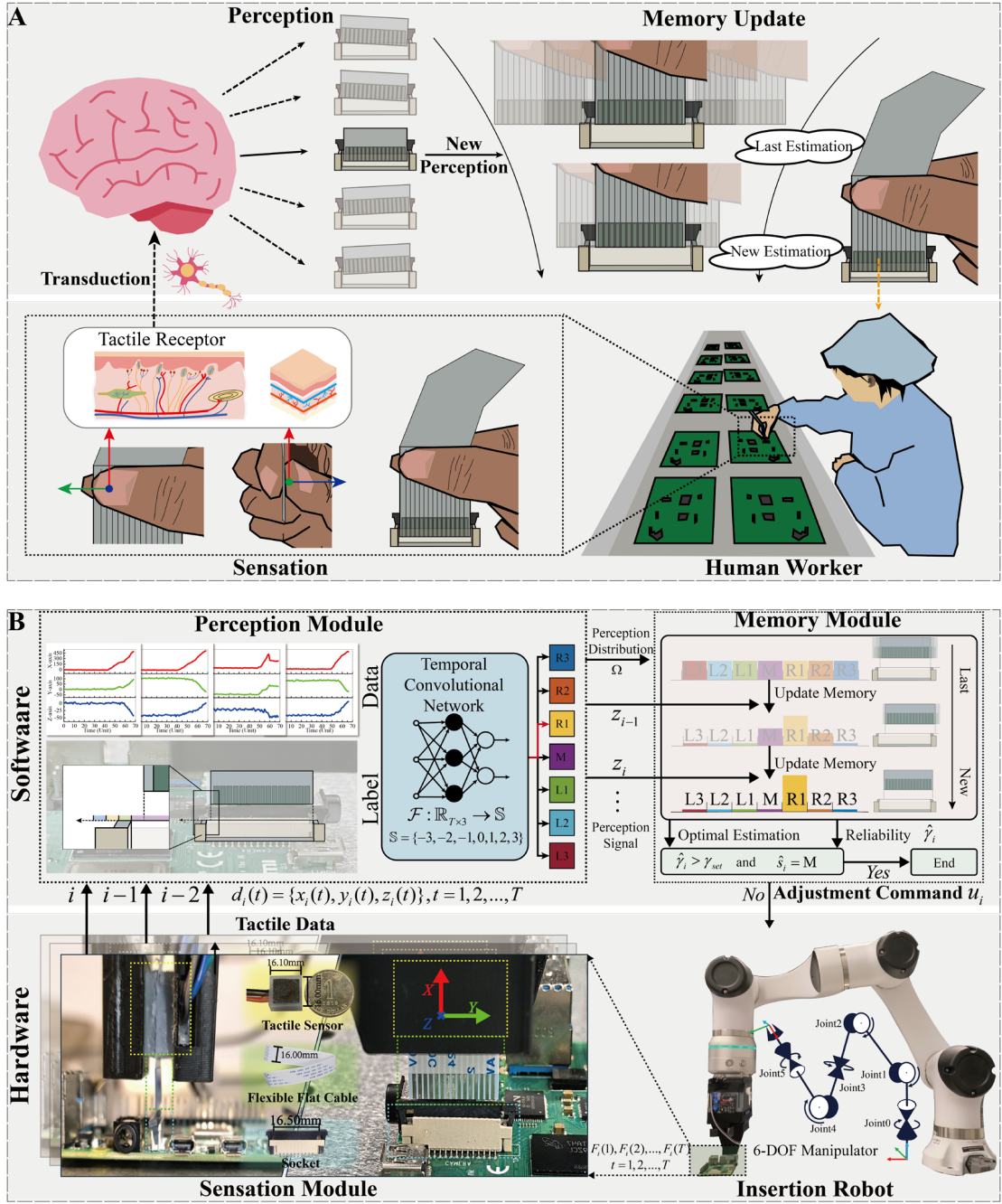

Fig. 1 An overview of the FFC insertion framework of the human and our robot. (A) is the mechanism of human insertion. The synergy of great sensing and intelligence, combined with the fusion of current perception and memory, facilitates reliable insertion. In the bottom-right corner, the human worker is inserting FFC, while his fingers with tactile receptors are shown next to him. The receptors sense both normal forces (blue arrow) and shear forces (red arrow and green arrow) which result from the contact between the FFC and socket. The force stimuli are transmitted to the brain via nerves and converted into perception signals that estimate the contact status. This signal helps update the potential states stored in memory. (B) presents the proposed highly reliable FFC insertion framework. In the bottom-right corner, a six-axis robotic arm functions as the system's motion component. Adjacent to it is the sensation module (see Section 2.2), which enables the robot to sense three-dimensional forces during insertion. The tactile sensory data are transmitted to the perception module (see Section 2.3), which acts as the brain, converting the sensory data into perception signals. The memory module (see Section 2.4) receives the perception signals of all iterations and uses them to update the reliability.

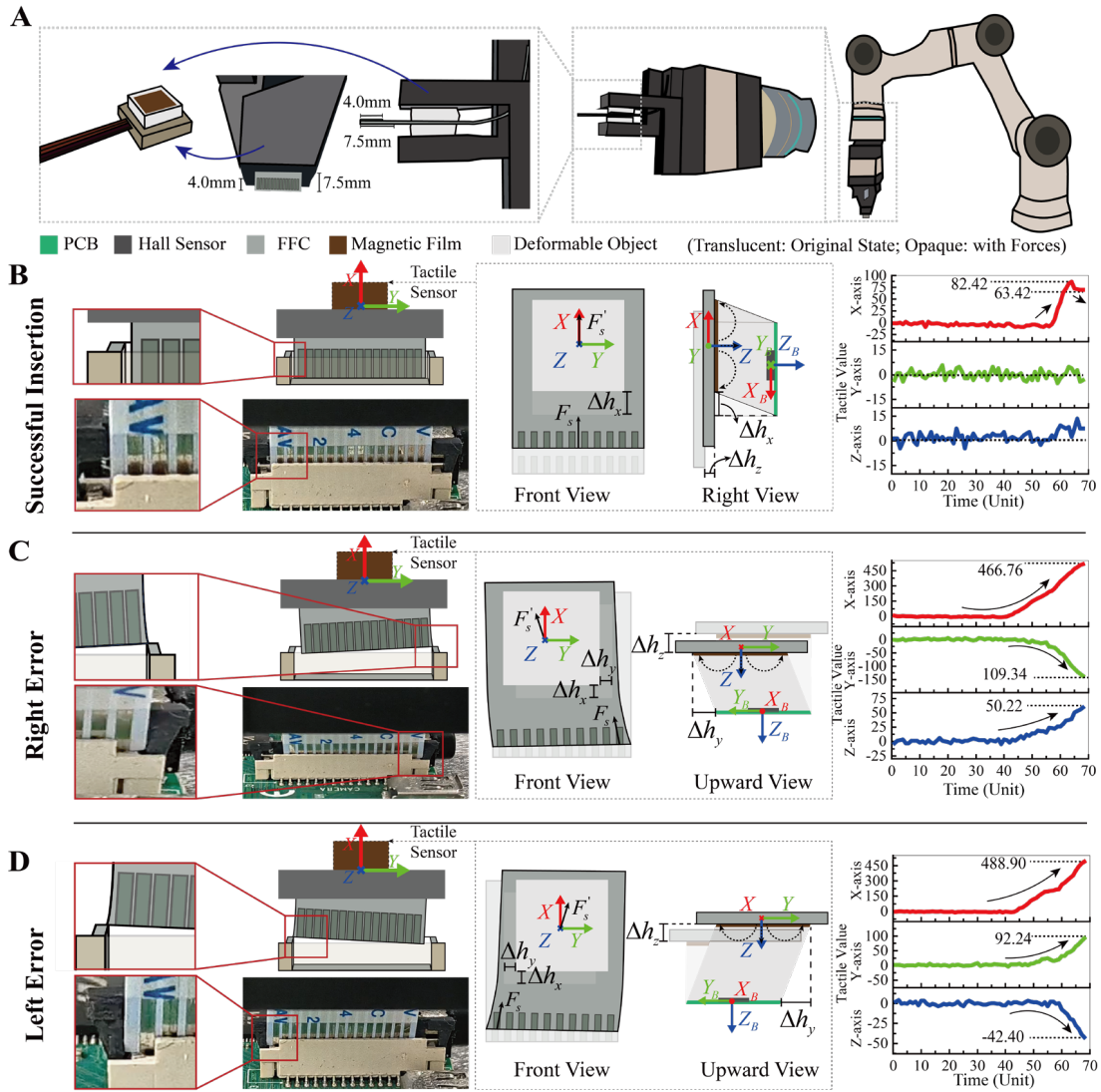

Fig. 2 The sensory module and the generated data from different insertion statuses. (A) The integration of sensation module. (B) Successful insertion of the FFC. (C) Failed insertion with right error. (D) Fail insertion with left error. Each case illustrates the contact status between the FFC and the socket, the influence on the sensor, and the corresponding tactile sensory data.

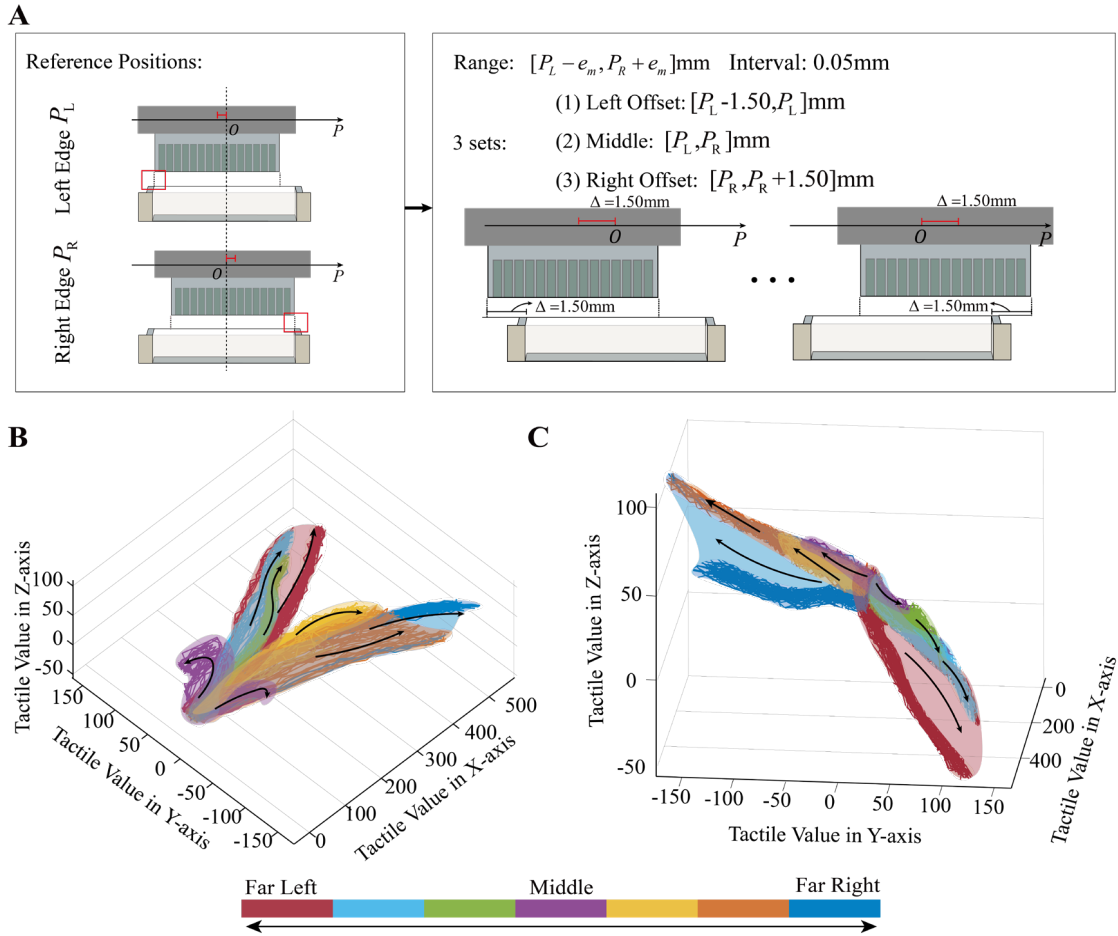

Fig. 3 The visualization of the tactile data of FFC insertion. (A) illustrates the process of data collection. (B) presents the three-dimensional tactile sensory data with different directions and magnitudes. The black arrows represent the direction of change in sensory data over time.

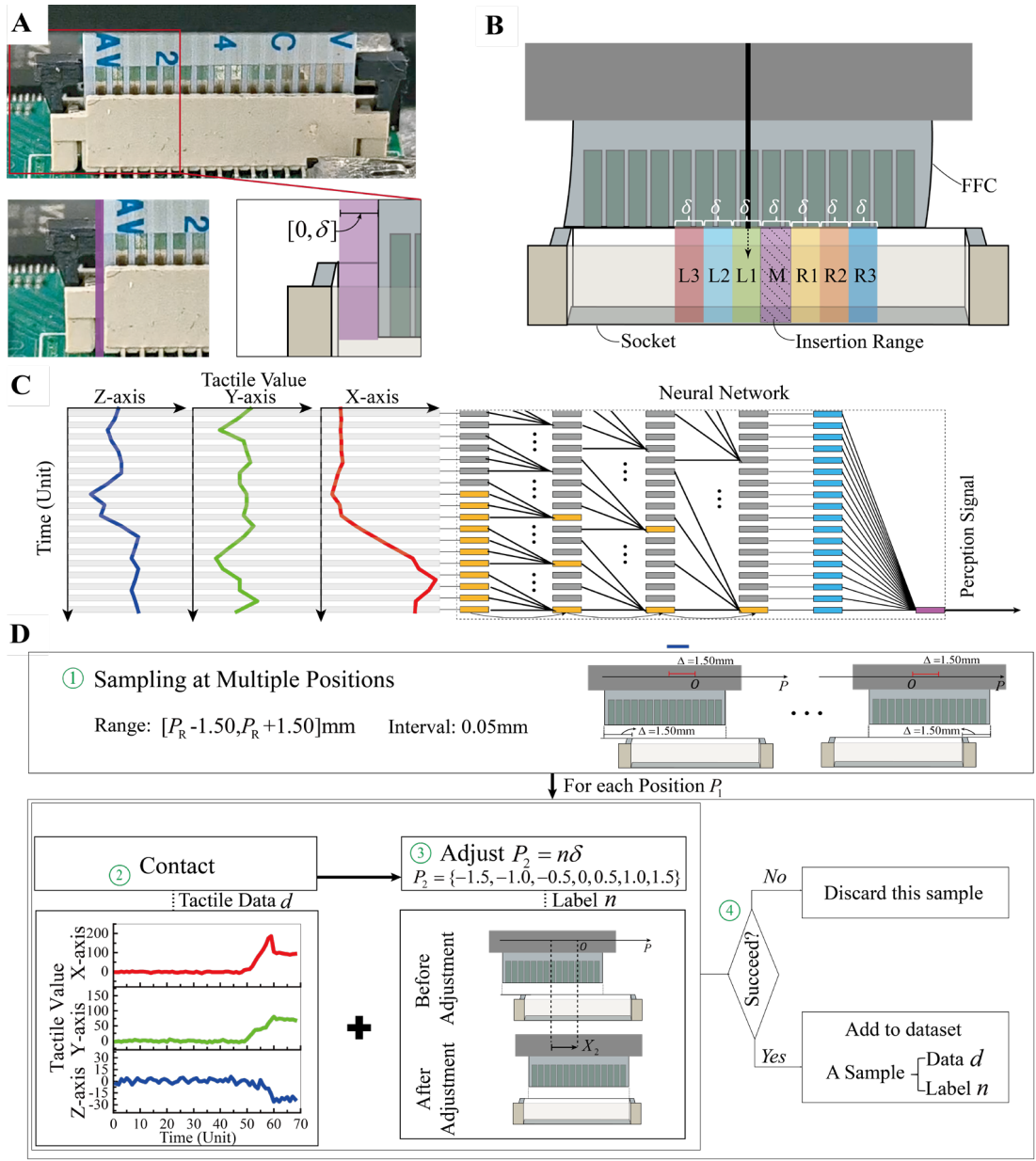

Fig. 4 Sketch of the construction of the perception module. (A) show the clearance between the FFC and the socket. (B) demonstrate the definition of the seven error regions. (C) illustrates the structure of the neural network for perception. (D) demonstrate the process of the dataset building for the perception model.

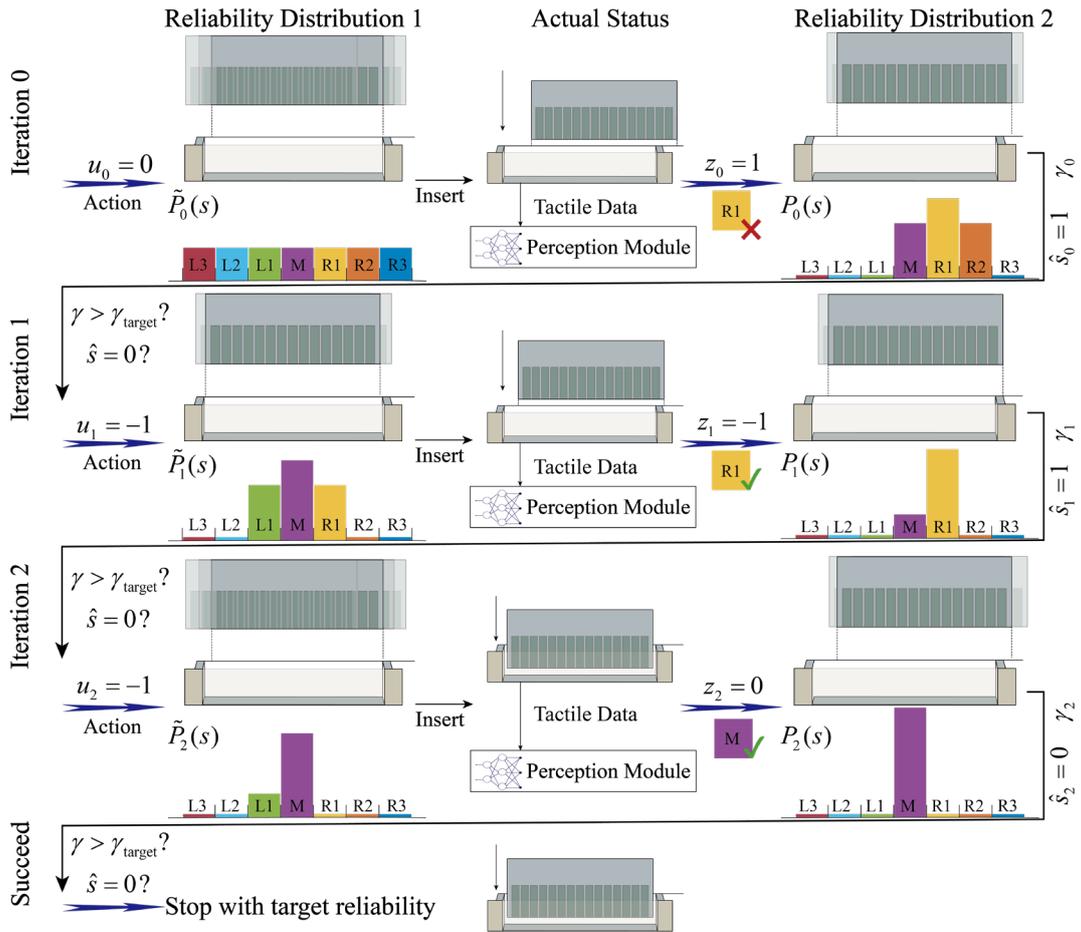

Fig. 5 Sketch of the FFC insertion with reliability control. The histogram represents the reliability distribution. The opaque FFC represents the status with the highest reliability, while the transparent FFC indicates statuses with lower reliability.

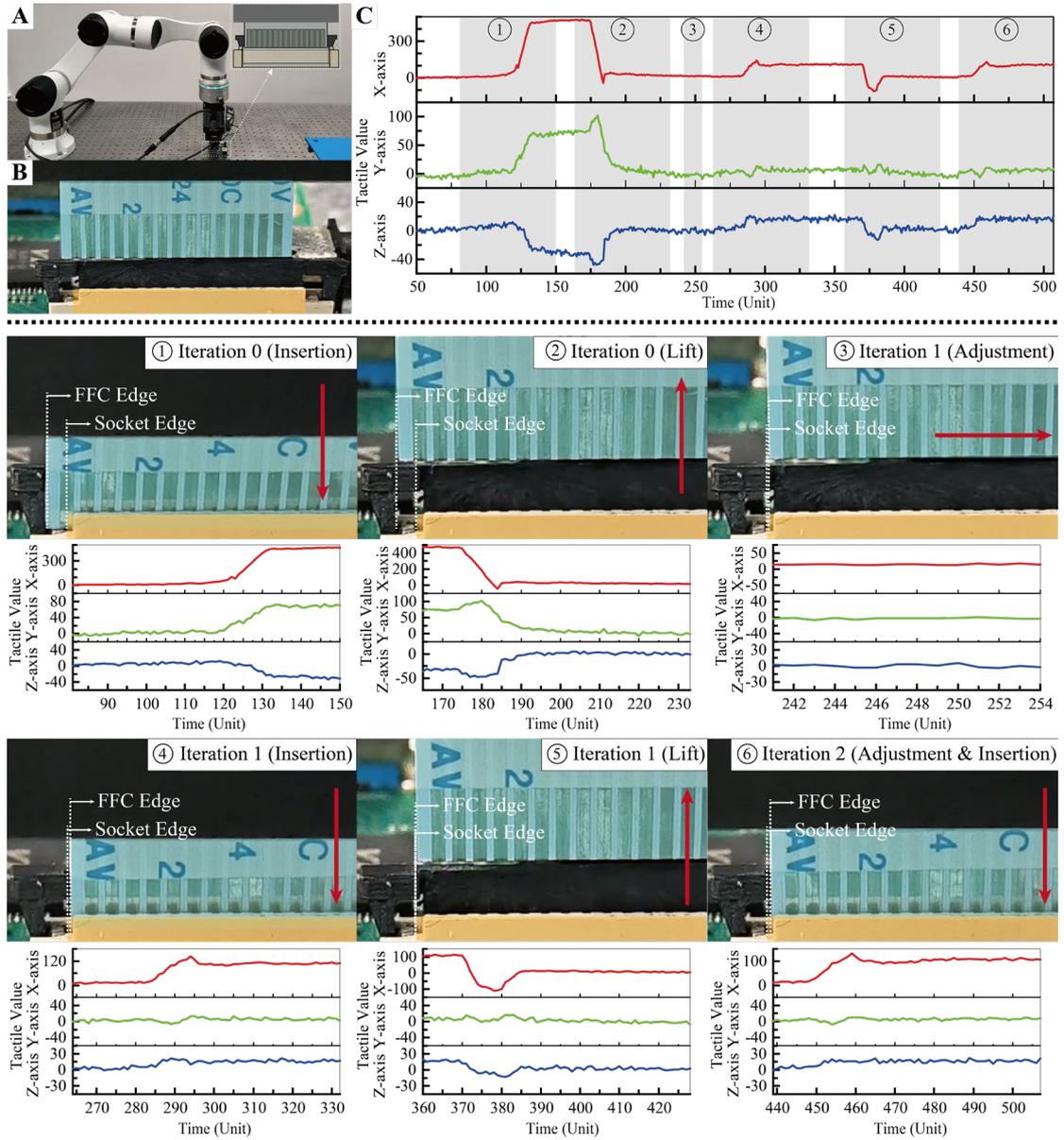

Fig. 6 The process and results of the FFC insertion tests. (A) shows the insertion setup, where the robot and the circuit board are mounted on a flat table. (B) presents the details of the FFC and the socket, depicting the status of the FFC with a left alignment error. (C) shows the tactile data of the process of FFC insertion, and the tactile data for each stage are displayed below. In these images, the red arrows indicate the direction of FFC movement, while the orange and blue blocks delineate the FFC and the socket respectively.

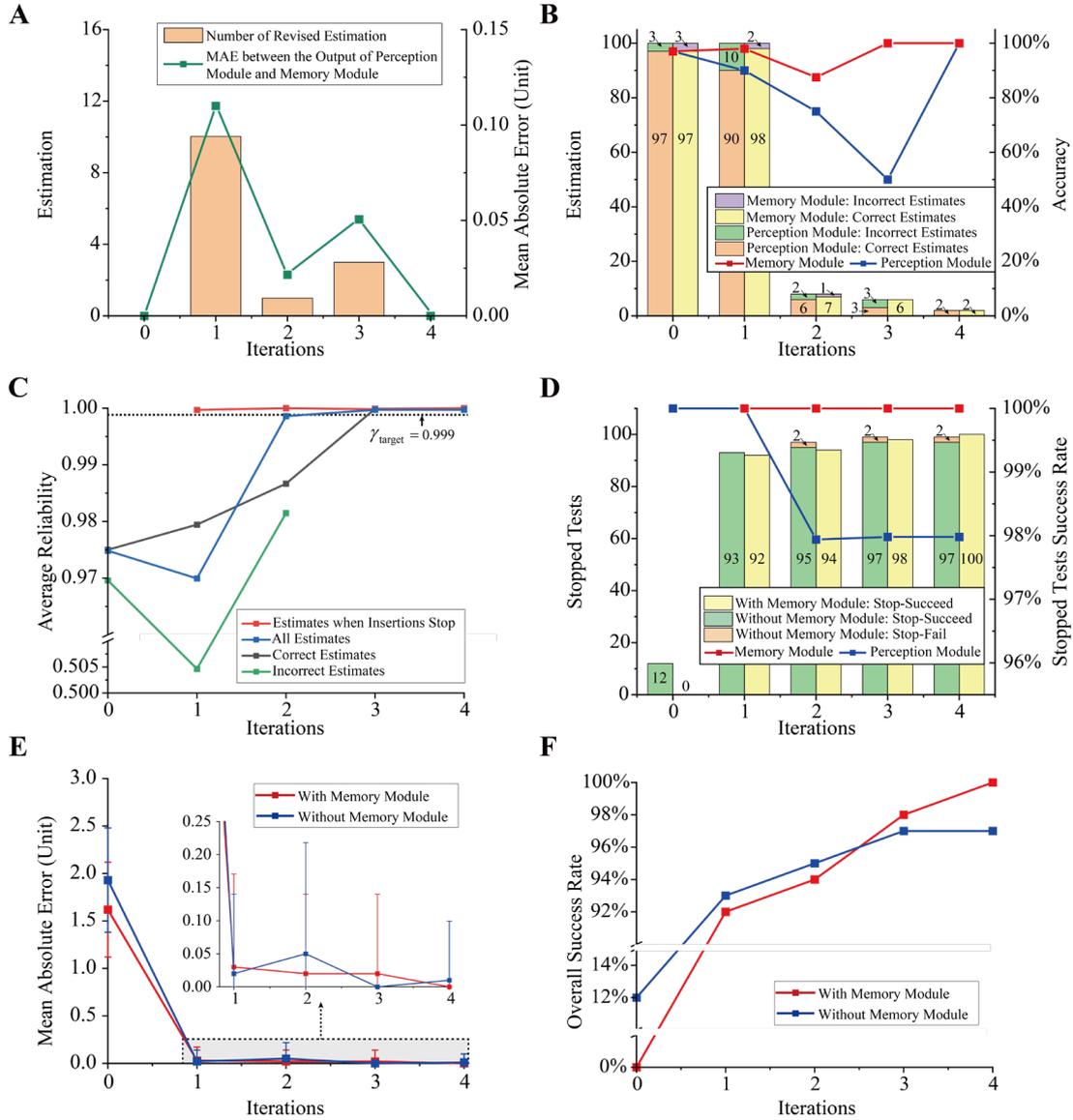

Fig. 7 The results of insertion tests for the solutions with and without the memory module. (A) presents the number of revisions and the average absolute revision values made by the memory module to the perception module. (B) presents the estimation accuracy and correctness of the perception module and the memory module. (C) shows the difference between the output of the perception module and the memory module in each iteration. (D) indicates the number of stopped tests and their success rate of the solutions with and without the memory module. (E) Presents the MAE (Mean of Absolute Error) between actual status and 0. (F) indicates the success rate with the iterations increases.

Table 1 The Precision of the Perception Module in Error Region Detection

Estimated Status

| | | L3 | L2 | L1 | M | R1 | R2 | R3 |
|---|---|---|---|---|---|---|---|---|
| Real Status | L3 | 98.18% | 1.82% | 0 | 0 | 0 | 0 | 0 |
| | L2 | 1.82% | 96.36% | 1.82% | 0 | 0 | 0 | 0 |
| | L1 | 0 | 0 | 100% | 0 | 0 | 0 | 0 |
| | M | 0 | 0 | 0 | 100% | 0 | 0 | 0 |
| | R1 | 0 | 0 | 0 | 1.82% | 98.18% | 0 | 0 |
| | R2 | 0 | 0 | 0 | 0 | 1.82% | 96.36% | 1.82% |
| | R3 | 0 | 0 | 0 | 0 | 0 | 3.63% | 96.36% |

Table 2 The Perception Distribution Matrix

Perception Signals

| | | L3 | L2 | L1 | M | R1 | R2 | R3 |
|---|---|---|---|---|---|---|---|---|
| Real Statues | L3 | 1 | 0 | 0 | 0 | 0 | 0 | 0 |
| | L2 | 0.0182 | 0.9636 | 0.0182 | 0 | 0 | 0 | 0 |
| | L1 | 0 | 0.0182 | 0.9818 | 0 | 0 | 0 | 0 |
| | M | 0 | 0 | 0.0182 | 0.9636 | 0.0182 | 0 | 0 |
| | R1 | 0 | 0 | 0 | 0.0182 | 0.9636 | 0.0182 | 0 |
| | R2 | 0 | 0 | 0 | 0 | 0.0189 | 0.9811 | 0 |
| | R3 | 0 | 0 | 0 | 0 | 0 | 0.0182 | 0.9818 |

## Acknowledgments

This work was partially supported by Hong Kong RGC General Research Fund (16203923) and Guangdong Basic and Applied Basic Research Foundation (GDST24EG02).

## Author Contributions

Z. L. designs the insertion framework, implements the system, leads the experiment, and writes the draft. X. Y. and D. G. design the fabrication method for the tactile sensor and offer suggestions for its integration. H. C. assists with implementing the network model in the perception module and engages in deep discussions about the concepts of the memory module. T. Z. and R. Z. provide suggestions for experiment design. Y. S. proposes the concepts and leads the project. All the authors read and revised the manuscript.

## Competing interests

The authors declare no competing interests.